\pdfoutput=1
\documentclass[11pt]{article}

\usepackage[final]{acl}
\usepackage{setspace}
\usepackage{amsmath}
\usepackage{enumitem}
\usepackage{color}
\usepackage{array}
\usepackage{threeparttable}
\usepackage{caption}
\usepackage{multirow}
\usepackage{makecell}
\usepackage{adjustbox}
\usepackage{booktabs}
\usepackage[skins,breakable]{tcolorbox}
\usepackage{times}
\usepackage{latexsym}
\usepackage{hyperref}
\usepackage{stfloats} 
\usepackage{lipsum,mwe,cuted}
\usepackage{float}

\usepackage[T1]{fontenc}
\usepackage[utf8]{inputenc}

\usepackage{microtype}

\usepackage{inconsolata}

\usepackage{graphicx}

\newcommand{\ours}{MetaTox}
\newcommand{\zhu}[1]{{\color{black}#1}}
\newcommand{\zyb}[1]{{\color{black}#1}}
\newcommand{\xu}[1]{{\color{black}#1}}
\newcommand{\newzyb}[1]{{\color{black}#1}}
\newcommand{\case}[1]{\emph{#1}}
\title{Enhancing LLM-based Hatred and Toxicity Detection with\\ Meta-Toxic Knowledge Graph}

\author{Yibo Zhao,
Jiapeng Zhu, 
Can Xu,
Yao Liu\textsuperscript{\rm 1}\thanks{Corresponding author: liuyao@cc.ecnu.edu.cn, xiangli@dase.ecnu.edu.cn}, 
    Xiang Li \textsuperscript{\rm 1}\footnotemark[1] \\
East China Normal University,
Shanghai, China \\
\\
}

\begin{document}
\maketitle

\begin{abstract}
The rapid growth of social media platforms has raised significant concerns regarding online content toxicity.
When Large Language Models (LLMs) are used for toxicity detection, two key challenges emerge: 
1) the absence of domain-specific \newzyb{toxicity} knowledge leads to false negatives; 2) the excessive sensitivity of LLMs to toxic speech results in false positives, limiting freedom of speech.
To address these issues, we propose a novel method called \emph{\ours}, leveraging graph search on a meta-toxic knowledge graph to enhance hatred and toxicity detection.
First, we construct a comprehensive meta-toxic knowledge graph by utilizing LLMs to extract toxic information through a three-step pipeline.
% , with toxic benchmark datasets serving as corpora. 
Second, we query the graph via retrieval and ranking processes to supplement accurate, relevant \newzyb{toxicity} knowledge.
Extensive experiments and case studies across multiple datasets demonstrate that our \emph{\ours}
% significantly decreases the false positive rate while boosting overall toxicity detection performance.
boosts overall toxicity detection performance, particularly in out-of-domain settings.
In addition, under in-domain scenarios, we surprisingly find that small language models are more 
competent. Our code is available at \url{https://github.com/YiboZhao624/MetaTox}.
%这里再加文本就会爆炸！

%While existing methods have shown promising performance, they are inherently constrained by training datasets that limits their generalization ability. To address these limitations, we propose a novel method called \emph{\ours} that leverages a meta-toxic knowledge graph, which is automatically extracted from toxic texts, to enhance toxicity detection capabilities. Our contribution is twofold: first, we construct a comprehensive toxic knowledge base called meta-toxic knowledge graph by employing LLMs to automatically extract the toxic information through a three-step pipeline including reasoning, self-checking extracting, and resolving. Second, we propose a graph query step to leverage the meta-toxic knowledge graph to improve the performance of LLMs on the downstream tasks, including toxic content detection and reasoning. Through extensive experimentation across multiple benchmarks, we demonstrate that our method significantly enhances both the performance and robustness of LLMs. Our code is available at \href{https://anonymous.4open.science/r/MetaTox-9070}{https://anonymous.4open.science/r/MetaTox-9070}
\end{abstract}

\textcolor{red}{\textbf{Disclaimer}: \textit{This paper describes toxic and discriminatory content that may be disturbing to some readers.}}

\section{Introduction}\label{sec:introduction}

Online social media platforms have become a major source of information 
% and communication tools 
for people worldwide. 
Meanwhile, 
they also provide a communication tool for spreading toxic content
including harassment, trolling, cyberbullying, and hate speech, which poses a serious and continual threat to the harmony of society~\cite{simpson2013dignity} and harms children's mental health~\cite{simpson2019won}. 
It thus becomes a critical task to detect the toxic content
%both explicit like ``{fuck you}'' and implicit like ``{How dark is my humor? It picks cotton}''
\zyb{, while cautiously avoiding hindering freedom of expression.}

\begin{figure}[t]
    \centering
    \includegraphics[width=0.95\linewidth]{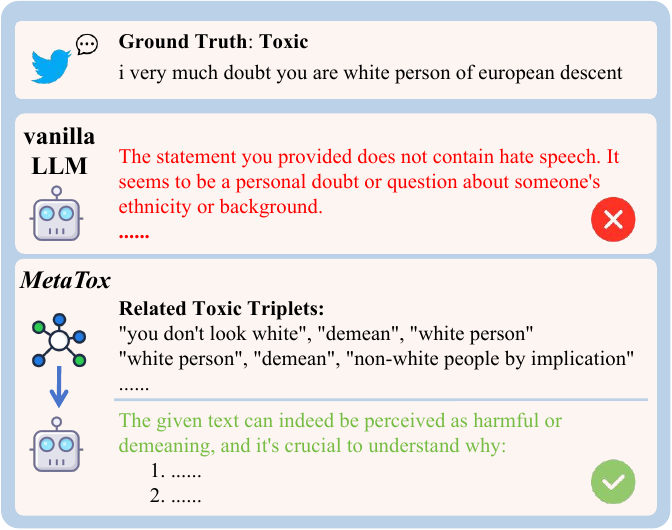}
    \caption[10pt]{A showcase of our method \emph{\ours}, which enhances the LLM to detect toxicity by injecting speech-related toxic knowledge in form of triplets.}
    \vspace{-3mm}
    \label{fig:introcase}
\end{figure}

To address the problem,
previous work can be mainly categorized into three types: (1) rule-based methods~\cite{chen2012detecting,gitari2015lexicon} relying on the pre-defined rules; (2) embedding-based methods~\cite{warner2012detecting, djuric2015hate, park2017one} leveraging text representations for classification; (3) transformer-based methods~\cite{bert_fastText, luu2021uit} 
employing transformer architectures with small-scale parameters. 
The first two types of approaches struggle with semantic understanding, limiting their ability to detect implicit toxicity. 
%\newzyb{The third category of methods utilizes small language models, which are widely applied due to their lightweight nature and ease of training.}
While methods of the third type can capture more complex semantics, 
their reliance on domain-specific training data results in significant performance degradation on out-of-domain data.
\newzyb{However, due to their lightweight nature and ease of training, they are widely applied.}

The rapid development of large language models (LLMs) has provided new insights into the detection of toxic contents.
\newzyb{Therefore, we try to explore whether LLM-based methods can outperform small models in this domain.}
Some studies~\cite{yang-etal-2023-hare, Zhang_Wu_Xu_Cao_Du_Psounis_2024} leverage LLMs as data augmentation tools to enhance toxicity detection. 
However, due to LLMs' lack of domain-specific knowledge in hatred and toxicity, the performance of these methods is still limited.
Further,
recent research~\cite{zhang-etal-2024-dont-go} highlights another issue: LLMs are extremely sensitive to groups or topics that may raise fairness concerns, such as race, gender, and religion, often leading to 
{false positive misjudgment}, \zyb{which impairs the freedom of speech}. For example, the benign post ``\case{forget this white nationalist mess. i'm america first...}'' is misjudged as toxic by LLM possibly because the model is too sensitive to the ``\case{white nationalist}''.

To supplement external domain knowledge for LLMs,
Retrieval Augmented Generation (RAG)~\cite{10.5555/3495724.3496517} has been widely adopted. 
% to complement domain-specific knowledge for LLMs. 
However, 
naive RAG-based methods heavily rely on the semantic similarity between the embeddings of queries and documents.
When query contents are implicitly toxic with vague semantics, retrieved documents could be irrelevant and impair the model performance.
Inspired by the recent success of GraphRAG~\cite{edge2024local}, \newzyb{as well as the inherent nature of toxic content targeting specific groups, regions, and other entities,} \zhu{we organize external knowledge into a meta-toxic knowledge graph, and dig out toxicity knowledge highly relevant to speech via graph retrieval. The retrieved triplets or paths, which represent concentrated toxicity knowledge derived from the originally fragmented toxic corpus, capture rich meta-toxic information. This process facilitates more accurate detection of toxicity, thereby alleviating the issue of false positive misjudgments.}
% if a meta-toxic knowledge graph is constructed,
% domain-specific knowledge from the graph can then be injected. 
% Furthermore,
% the semantics of implicit toxicity can be captured by performing knowledge graph retrieval.
% The retrieved triples contain rich meta-toxic information and can be used to alleviate 
% In this way,
% the false positive misjudgment problem. 
% can be inherently alleviated.

% However, the RAG-enhanced models rely heavily on the quality of the semantical embedding and the retrieval quality, which means when the toxic content conveys the implicit meaning, the queried documents may be irrelevant and impair the performance of the RAG-enhanced models. 

% However, if our knowledge base is graph-structured, 
% we can map the entities in the query to the graph and retrieve the related knowledge to achieve a more precise information retrieval.

% At the same time, by introducing
% knowledge graph,
% the graph structure, 
% we can also improve the multi-step reasoning ability by path search in the graph, 
% which alleviates the second problem mentioned above.

% Therefore, 
In this paper, we propose \emph{\ours},
which utilizes a meta-toxic knowledge graph and LLMs for hate and toxic speech detection.
% detection with meta-toxic knowledge graph.
% Specifically,
% a meta-toxic knowledge graph?
We first construct a meta-toxic knowledge graph from existing toxic benchmark datasets
by designing a three-step pipeline: \emph{rationale reasoning}, \emph{triplet extraction}, and \newzyb{\emph{entity resolution}}.
Specifically,
% which contains the hatred and toxicity domain knowledge.
rationale reasoning aims to reason about what contents trigger toxicity in the speech;
% to provide a more explicit understanding of the toxic content.
% alleviate the implicit meaning problem. 
triplet extraction involves extracting toxic entities and relations, with a self-checking strategy \zhu{to ensure the quality of these triplets;}
% to ensure the triplets are correctly formatted and toxic; 
% mitigating the hallucination; 
and \newzyb{entity resolution} merges nodes and relations with similar semantics respectively.
% a novel method to complement the hatred and toxicity domain knowledge for LLMs by constructing Meta-toxic Knowledge Graph, which only contains the hatred and toxicity domain knowledge extracted from the existing datasets. 
Subsequently, given a downstream potentially toxic speech, we query the knowledge graph by \emph{retrieval} and \emph{ranking} for toxicity detection.
\zyb{Retrieval returns relevant paths to the speech, while ranking further filters noise and refines the triplets in paths.}
% what about triplets??
% we retrieve the knowledge graph for relevant toxic information, which is further denoised by ranking.
We then leverage the extracted triplets that serve as toxic prompts to boost the LLMs' capability in toxicity detection. \newzyb{\textbf{The entire process is training-free}.} A showcase of our method is depicted in Figure~\ref{fig:introcase}. Our contributions are summarized as follows.
% in the downstream tasks including classification and reasoning.

% For knowledge graph construction,
% % construction step, 
% we design a three-step pipeline including \emph{rationale reasoning}, \emph{triplet extraction}, and \emph{duplicate removal}.
% % which constructs a reliable and robust knowledge graph. 
% Specifically,
% rationale reasoning aims to reason about what contents trigger the toxicity in the speech;
% % to provide a more explicit understanding of the toxic content.
% % alleviate the implicit meaning problem. 
% triplet extraction is to extract toxic entities and relations with a self-checking strategy to check if the triplets are correctly formatted and toxic; 
% % mitigating the hallucination; 
% duplicate removal merges nodes and relations with similar semantics, respectively. 
% and separately merge relationships with similar meanings, 
% which reduces the size of the graph and improves the query efficiency.

% For knowledge graph query, 
% we employ a retrieval-reranking method to provide concise and related knowledge for LLMs. 
% The reranking step is designed to denoising by filtering out the irrelevant knowledge and prioritizing the most relevant knowledge for LLMs. Finally, we format the triplets by sequentially concatenating the head entity, relation, and tail entity to serve as complementary knowledge for LLMs.

\begin{itemize}[leftmargin=*, itemsep=0pt, parsep=0pt, topsep=0pt, partopsep=0pt]
\item We construct a novel meta-toxic knowledge graph. To our best knowledge, it is the first domain-specific knowledge graph for hate and toxic speech detection. We will open-source the knowledge graph upon paper acceptance.
 
\item 
 We propose an effective \newzyb{\textbf{training-free}} method \emph{\ours} for hate and toxic speech detection.
 % based on LLMs and meta-toxic knowledge graph.
 In particular, \emph{\ours} addresses the notorious false positive misjudgment issue, decreasing the ethical risk of hurting the freedom of speech.
 % in most existing LLM-based methods.
 
 % We first propose a new paradigm leveraging graph-structured knowledge base to enhance the LLMs' ability in the toxic content detection tasks, which complements the domain-specific knowledge and clarifies the reasoning process of the LLMs.
 
 % and query the graph to enhance the LLMs' ability. 
 % We introduce the self-checking and resolving mechanism to promise the quality of the graph and leverage the filtering and reranking to permit precise information retrieval.
 
 \item We conduct extensive experiments to evaluate the performance of \ours.
 % our proposed method.
 % verify the effectiveness and robustness of our proposed method compared to the vanilla LLMs, RAG-enhanced approaches, \newzyb{and supervised learning methods}. \newzyb{}
 % \item We 
 Surprisingly, we find that small language models are more suitable for in-domain toxicity detection, while our method is superior in out-of-domain settings.
 This further sheds light on model selection in various toxicity detection scenarios.
 
\end{itemize}

\section{Related Work}\label{sec:relatedwork}

% \subsection{Hatred and Toxicity Detection}

% Since the advent of social media, the detection of hatred and toxicity has become a pivotal research area. 
\xu{Previous methods can mainly be classified into rule-based, embedding-based, and transformer-based, and LLM-based approaches.}
% Hateful and toxic speech, as defined by~\cite{dixon2018measuring}, refers to ``rude, disrespectful, or unreasonable language likely to drive individuals away from a discussion.'' This encompasses a wide range of harmful content, including hate speech, harassment, offensive language, biases, and other detrimental expressions.

% Since the rise of social media, toxicity detection has emerged as a critical topic. According to~\cite{dixon2018measuring}, toxic speech encompasses ``rude, disrespectful or unreasonable language that is likely to make someone leave a discussion'', covering a broad spectrum of contents including hate speech, harassment, offensive language, bias, and other harmful expressions.

Rule-based methods~\cite{chen2012detecting, gitari2015lexicon} rely on predefined rules that match specific patterns within the text. \cite{liu2015new} uses the insults and swears words to form a dictionary for hatred and toxicity detection.
Embedding-based approaches, on the other hand, \xu{combine text representations obtained by methods like word2vec~\cite{djuric2015hate, park2017one}.}
% such as Part-of-Speech (POS) tagging~\cite{warner2012detecting}, , and traditional machine learning algorithms. 
Transformer-based methods~\cite{bert_fastText, luu2021uit}, \xu{typically perform fine-tuning on pre-trained models like BERT.} 
% are constrained by the domain-specific nature of their training data. 
Recent studies have explored the use of  LLMs, \xu{which typically serve as a powerful data augmentation tool. \newzyb{Some methods~\cite{lee-etal-2024-exploring-cross} directly prompt LLMs to generate explanatory information about the text, which is then combined with the original text to train a detection model. Furthermore, some approaches~\cite{yang-etal-2023-hare, Zhang_Wu_Xu_Cao_Du_Psounis_2024} employ structured reasoning methods such as Chain-of-Thought (CoT)~\cite{10.5555/3600270.3602070} and Tree-of-Thought (ToT)~\cite{10.5555/3666122.3666639} to infer hidden semantics of input text.}
{Some studies aim to enhance the robustness of hatred and toxicity detection. For example, TextAttack~\cite{attack1} improves robustness through data augmentation, while~\citet{attack2} proposes the ToxicTrap modular, leveraging adversarial training to strengthen model resilience.}

\newzyb{However, previous approaches have either overlooked related real-world knowledge or relied solely on distilling the limited internal knowledge of LLMs. In contrast, our method is the first to build a hate-related KG and introduce it into hatred and toxicity detection, offering interpretive knowledge injection and multi-hop inference ability.}
}

\section{Methodology}

In this section, \zhu{we present \emph{\ours}, our proposed method for enhancing hatred and toxicity detection using a meta-toxic knowledge graph. \emph{\ours} consists of two stages: (1) construction of the meta-toxic knowledge graph through a three-step pipeline, and (2) querying the graph to enhance the downstream task of binary classification for toxicity detection. We begin with the data collection process, which lays the foundation for constructing the knowledge graph.}

% In this section, we present our method, \emph{\ours}, which consists of two main steps: (1) constructing the meta-toxic knowledge graph through a three-step pipeline, and (2) querying the graph to enhance the downstream task, binary classification. The process of constructing the meta-toxic knowledge graph begins with data collection as the foundational step.

% In this section, we present our method: \emph{\ours} with two steps: (1) Construct the meta-toxic knowledge graph with a three-step pipeline, (2) query on the graph to enhance downstream tasks, including binary classification and reasoning. To construct the meta-toxic knowledge graph, we start with data collection at the very beginning.

\subsection{Data Collection}\label{subsec:data-collection}

For meta-toxic knowledge graph construction, we leverage three well-established English datasets: HateXplain~\cite{mathew2021hatexplain}, ToxicSpans~\cite{pavlopoulos2021semeval}, and IHC~\cite{elsherief-etal-2021-latent}.
\zhu{Since our goal is to enhance toxicity detection by supplementing the LLM with accurate and curated toxicity knowledge, the knowledge graph should serve as a domain-specific corpus that reflects only toxic content. Therefore, we exclusively retain toxic samples labeled as ``toxic'', ``hate'' or ``offensive'' from the training sets of these datasets to construct the knowledge graph.

For toxicity detection, we use the test sets from the three datasets, aligning each sample's label with either ``toxic'' or ``non-toxic'' to formalize a binary classification task. This enables us to evaluate the LLM's performance improvement in detecting potentially toxic speech with the support of our meta-toxic knowledge graph.}
% From these sources, we carefully extract and filter toxic content samples from the training set.

% To evaluate the performance of \emph{\ours}, we use the test sets from the same datasets, ensuring consistency in the assessment of our approach. Furthermore, to demonstrate the robustness and generalization of our approach, we specifically investigate how the meta-toxic knowledge graph constructed from Toxicspans enhances the performance when applied to the HateXplain and the IHC datasets.

% For our specific task, we apply a series of preprocessing steps to filter the data. In the training set, we retain only entries labeled as ``Toxic'' or other similar labels such as ``hate'', ``offensive'' and others, to construct the meta-toxic knowledge graph. For the test sets, we align different datasets to the toxic and non-toxic labels.

\subsection{Graph Construction}

\begin{figure}[!t]
    \centering
    % \vspace{0.3cm}
    \includegraphics[width=0.9\linewidth]{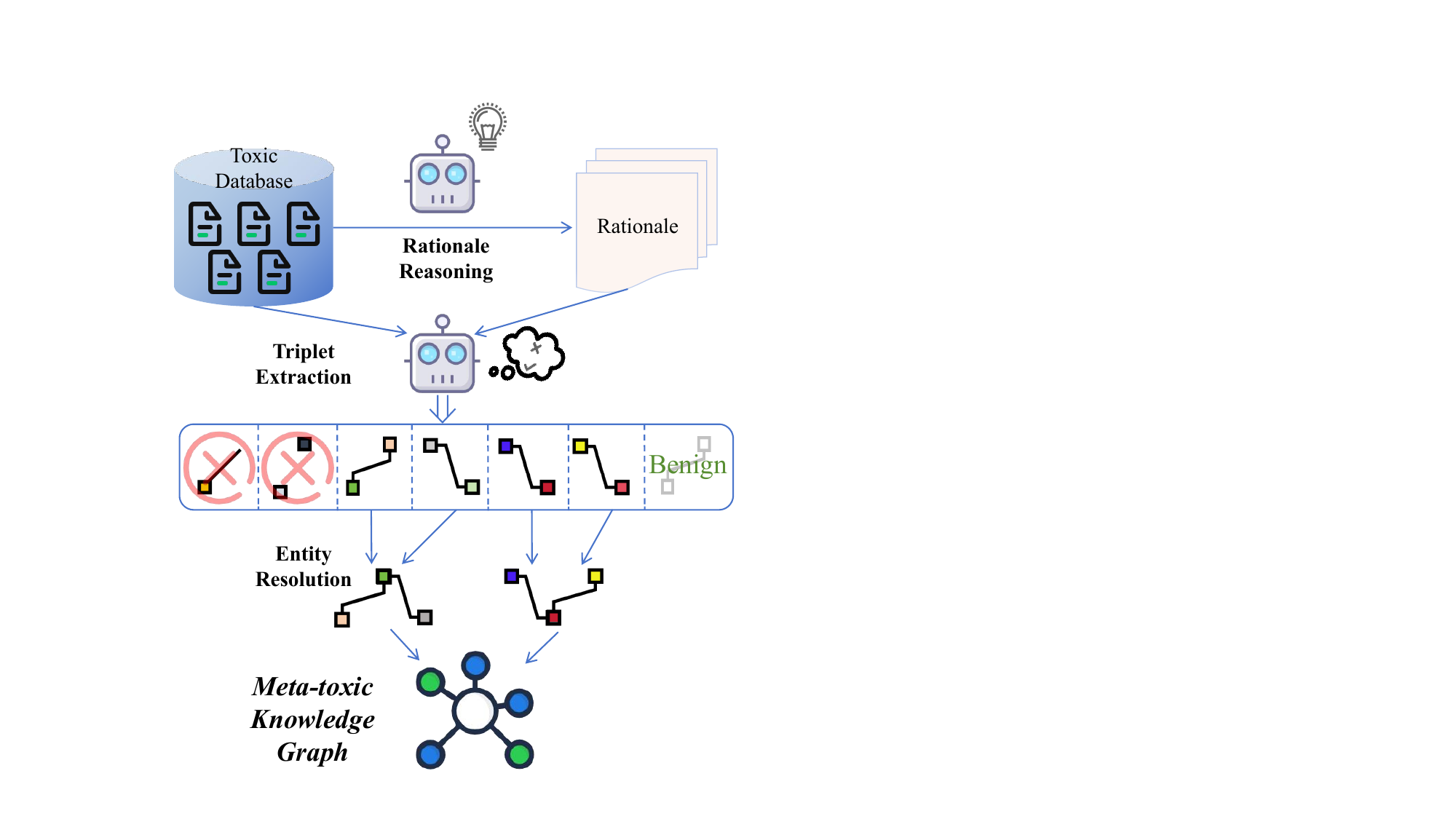}
    \caption[10pt]{The pipeline of meta-toxic knowledge graph construction, including rationale reasoning, triplet extraction, and \newzyb{entity resolution}.}
    \vspace{-0.2cm}
    \label{fig:graphConstructionPipeline}
\end{figure}

% To strike a balance between flexibility and efficiency, w
We propose a three-step process to build the meta-toxic knowledge graph \zhu{based on solely toxic samples}: 
(1) \emph{rationale reasoning}, which involves identifying the contents that trigger toxicity in speech, thereby uncovering implicit toxic meanings; 
(2) \emph{triplet extraction}, where toxic entities and relations are extracted using a self-checking mechanism to ensure the triplets are correctly formatted and toxic; 
(3) \emph{\newzyb{entity resolution}}, which merges semantically similar entities and relations to reduce noise and shrink the graph. 
The three steps are illustrated in Figure~\ref{fig:graphConstructionPipeline} and described in detail below.

\subsubsection{Rationale Reasoning}

\zhu{To construct the meta-toxic knowledge graph from initial toxic speech, it is essential to first identify the core toxic elements, i.e. entities and relations, within the speech. However, toxic semantics are often implicit and abstract, involving concepts such as race, gender and religion, rather than specific named entities. This introduces challenges in directly extracting these elements in a single step.}
% In this step, we take the toxic speech as input and apply a LLM to generate rationales for further processing.
% However, identifying toxic content presents unique challenges, as the toxic meaning is often implicit and directed at demographic groups -- such as race, gender, and religion -- rather than specific named entities. This makes direct extraction of toxic elements particularly challenging. 

To address this, we \zhu{employ a rationale reasoning step prior to triplet extraction, which applies LLMs to articulate why the speech is considered toxic, with relevant elements leading to this conclusion naturally incorporated into the rationale.
For example, by informing the LLM that ``\case{white lives matter event}'' belongs to hate speech, the LLM explains: ``\case{a counter-movement to the Black Lives Matter movement, shows potential harm to the Black Lives Matter movement}''. This explanation not only provides the context that triggers toxicity but also highlights the toxic entities involved.
Hence, this rationale reasoning step can be interpreted as a form of data augmentation, which draws out implicit toxic elements from the speech and makes them more explicit to better guide the LLM’s reasoning logic towards toxicity, thereby reducing difficulties for subsequent triplets extracting.}
% To address this, we implement a reasoning step that explicitly articulates why specific content is classified as toxic and which elements are related to the toxic meaning. This approach makes the targets and their relationships more clearly identifiable. 
% For example, the content ``white lives matter event'' is explained as ``a counter-movement to the Black Lives Matter movement, shows potential harm to the Black Lives Matter movement'', which not only provides context triggering toxicity but also highlights the explicitly related entities.
We leverage in-context learning to enhance LLM's ability to generate high-quality rationales, using carefully designed prompts. A detailed description of the prompt can be found in Appendix~\ref{appendix:reasoning-prompt}.

\newzyb{It is important to emphasize that while LLMs exhibit excessive sensitivity and struggle to correctly judge when used for toxic content detection, our approach explicitly informs the model that the given text is toxic and leverages the model for reasoning rather than direct judgment or generation. This equates to \textbf{providing the model with prior knowledge, where the model only needs to deduce the reasons for the toxicity from the text.}} {A similar approach has also been adopted and validated in studies such as STaR~\cite{star}, which guide models in generating new rationales based on correct answers, thereby helping them learn to solve increasingly difficult problems using the generated rationales.}

% While some approaches discussed in Section \ref{sec:relatedwork} utilize CoT reasoning with LLMs, research by \cite{NEURIPS2022_9d560961} indicates that the CoT's effectiveness is primarily limited to large-scale models, with significant performance degradation observed in smaller models such as PaLM 62B. Given that our implementation utilizes models of 7B and 14B parameters, the CoT approach proves suboptimal for our situation, but it remains a viable option when greater computational resources are available.

\subsubsection{Triplet Extraction}\label{subsec:triplets-extracting}
\begin{figure*}[t]
    \centering
    % \vspace{0.3cm}
    \includegraphics[width=0.9\linewidth]{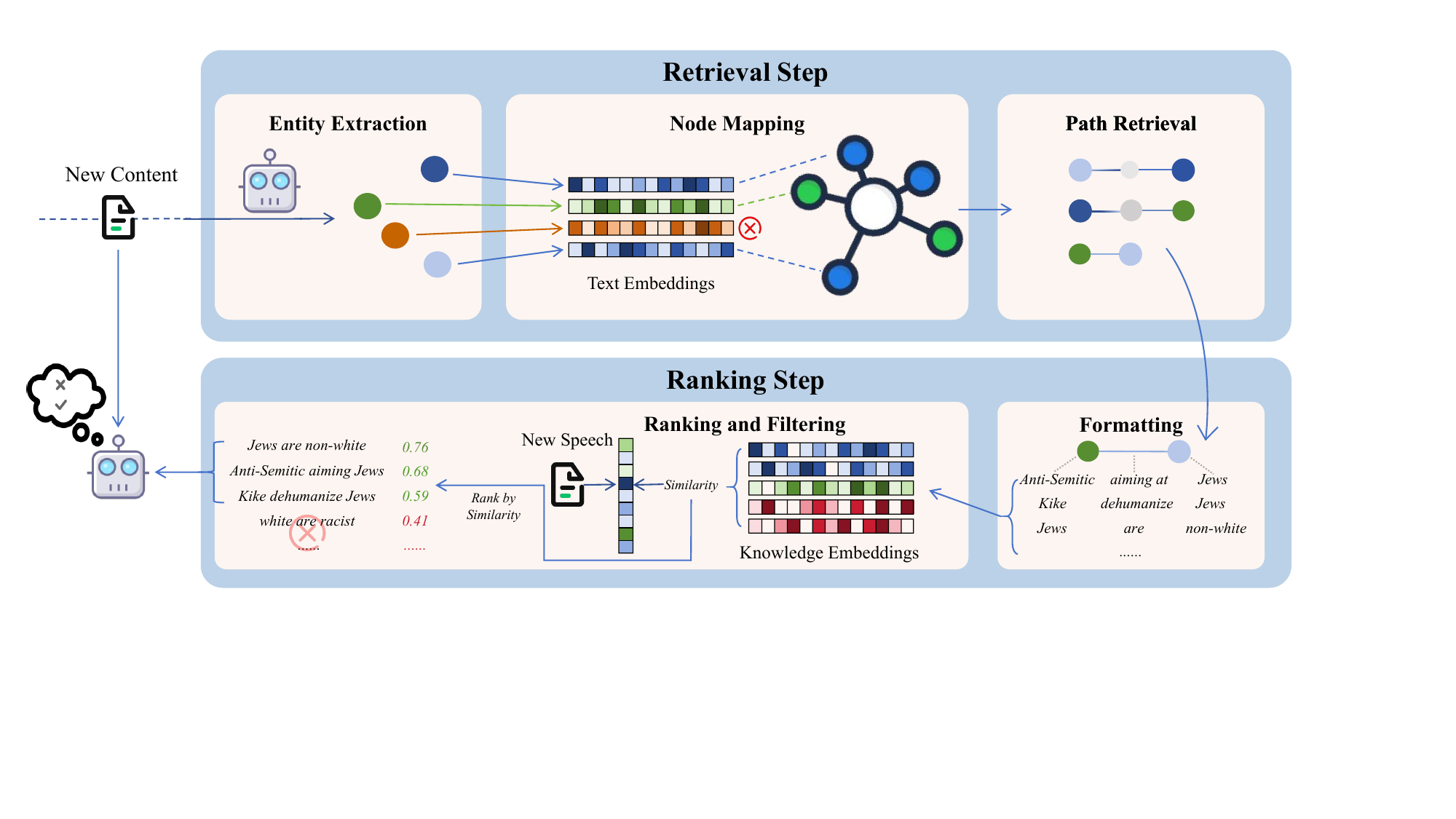}
    \vspace{0cm}
    \caption[10pt]{Graph query pipeline of \emph{\ours}. We propose five steps including entity extraction, node mapping, path retrieval, formatting, and ranking and filtering to inject accurate knowledge to the LLM.}
    % \vspace{-0.2cm}
    \label{fig:graphQueryPipeline}
\end{figure*}

In this step, 
% we take the toxic content and the reasoning results as input to extract the triplets. 
\zhu{we take the toxic speech and the corresponding rationales as input and prompt the LLM to extract toxic triplets,} represented in the Subject-Predicate-Object (SPO) format, such as ``{(\case{white lives matter, is against, black lives matter})}''.
\zhu{Specifically, we implement a template that instructs the LLM to extract triplets triggering hatred, given two exemplars to guide the LLM in understanding the extraction process and outputting triplets in a unifying SPO format.}
{It is worth noting that the extracted relations are not constrained to a predefined set but are determined by the LLM. Specifically, we include a few illustrative examples of potential relations in the prompt template and instruct the LLM to generate triplets using relations that best reflect the underlying semantics, rather than selecting from a fixed list.}
The details of the prompts are shown in Appendix~\ref{appendix:extracting-prompt}.

% Unlike conventional knowledge graph construction, our task presents two challenges: (1) the absence of labeled data and predefined relations, and (2) the ambiguous nature of
% the content, 
% \zhu{toxic semantics,}
% which often includes slurs and abbreviations. 
% These challenges make existing methods, which primarily focus on extracting predefined relations from high-quality corpora, less effective for extracting toxic triplets.
% Existing methods that primarily focus on extracting predefined relations from high-quality corpora \zhu{may degenerate under our scenario.}

% To address these challenges, we adopt an approach that leverages LLMs, enhanced with additional self-checking steps. 
% First, LLMs are used to automatically extract the toxic triplets and return them in a Subject-Predicate-Object (SPO) format, such as ``(white lives matter, is against, black lives matter)''. 
% To improve accuracy, we implement a few-shot prompt that defines the SPO format, provides some examples of possible relations, and demonstrates how to extract the triplets.
% The detailed prompt engineering is presented in Appendix \ref{appendix:extracting-prompt}.

% This approach capitalizes on the LLMs' comprehensive semantic understanding, which helps address the formal challenge in triplet extraction.

However, we observed two main issues with the extracted triplets.
First, due to occasional imprecision in following instructions, the LLM may generate triplets that deviate from the standard SPO format, like omitting a subject or object.
Second, the triplets may fail to capture toxic semantics, possibly due to inherent ambiguity in hate speech.

\zhu{Thus, we propose a self-checking mechanism to refine the extracted triplets, as illustrated with pictorial cases in Figure~\ref{fig:graphConstructionPipeline}. For triplets that are incorrectly formatted, we discard them using regular expressions. For triplets that fail to capture toxic semantics, we urge the LLM to filter out non-toxic triplets with few-shot prompting, ensuring that only those triplets capable of evoking hatred and toxicity are retained.} Details of the filtering prompt are provided in Appendix~\ref{appendix:filtering-prompt}.

% To mitigate these problems, we propose a two-fold solution. For the format consistency issue, we apply a rule-based post-processing method utilizing regular expressions to discard triplets that deviate from the correct format. For the second issue, similar to the preceding section, we use LLMs with few-shot prompting to evaluate whether the extracted triplets accurately capture the toxic content. The prompt specifies the correct output format, explains how to correct misformatted triplets, and provides examples for evaluating the triplets. The details of this prompt are documented in Appendix \ref{appendix:filtering-prompt}.

\zhu{After applying the self-checking mechanism, we obtain a curated set of toxic triplets that explicitly reflect toxic semantics, effectively leveraging external domain knowledge.}

\subsubsection{\newzyb{Entity Resolution}}

In this step, we take the extracted triplet elements (entities and relations) as input, applying a clustering algorithm to merge these elements and ultimately generate a meta-toxic knowledge graph with \newzyb{entities resolved}. Unlike traditional entity resolution methods for knowledge graphs, our approach simplifies the standardization process by assigning identical names to entities within the same cluster, rather than relying on specific Internationalized Resource Identifiers (IRIs).
% Specifically, in this way, we removed duplicated entities and relations, therefore decreasing the graph size and optimizing query efficiency.

\zhu{To perform clustering on the triplet elements}, we first use a pre-trained BERT model~\cite{devlin-etal-2019-bert} to encode the textual attributes of entities and relations into embeddings. 
% In terms of element representation, we utilize the pre-trained BERT model~\cite{devlin-etal-2019-bert} to independently encode the textual attributes of entities and relations.
% While some methods~\cite{garciaduran2018kblrnendtoendlearning, 10.1007/978-3-319-93417-4_38} incorporate graph structural information and use Graph Neural Networks (GNNs) to enhance representations, we find it is unnecessary for our task. This is because, during the query process, we embed the query text using BERT and already explicitly incorporate structural information via graph path search. If we were to use GNNs for node representation, the embeddings of the nodes and the query would no longer be aligned in the same space, and the connected nodes might not be semantically related -- potentially even contradictory, as seen in the case of ``white lives matter'' and ``black lives matter''. For these reasons, we opt not to use GNNs for node representation.
\zhu{We then apply a clustering algorithm to group similar entities and relations respectively, based on their textual embeddings. For deduplication, we determine the name that appears most frequently within each cluster and assign it as the unified name for all elements in that cluster. In other words, we use clustering combined with a voting scheme to merge similar elements.}
% This process employs the Agglomerative Clustering algorithm, which groups similar entities and relations relatively based on their embeddings. Within each cluster, the most frequently selected element name is chosen to represent all elements in that group.
This method not only helps resolve slight spelling variations, such as ``\case{Jew}'' and ``\case{jew}'', but also standardizes elements with different names referring to the same concept, such as ``\case{LGBT}'' and ``\case{LGBTQ+}''. To prevent over-merging, we set a relatively high similarity threshold.
% This approach not only helps deduplicate elements with slight spelling variations, such as ``Jew'' and ``jew'', but also standardizes elements with different names referring to the same concept, such as ``LGBT'' and ``LGBTQ+''. To prevent over-merging, we set a relatively high similarity threshold.

\zhu{The benefits of \newzyb{entity resolution} are twofold. On the one hand, it reduces noise by eliminating unnecessary spelling differences and retaining representative expressions supported by most elements. On the other hand, it decreases the overall size of the graph, which in turn enhances the efficiency of graph retrieval.}

% \zhu{TODO: Naive RAG methods, particularly based on dense retrieval, construct indexing independently among paragraphs or chunks embeddings, while graph-based RAG organizes knowledge with similar semantics by acting as neighboring nodes.}

% , we both deduplicate the elements with slightly different spelling like ``Jew'' and ``jew'', and resolve the entities indicating the same element with different names to a unified name like 
\subsection{Graph Query}
\zhu{Given a potentially toxic speech, we devote to querying on the meta-toxic knowledge graph to supplement speech-related toxic triplets, thereby providing the LLM with accurate, domain-specific guidance when judging the hatred and toxicity of the given speech.} 
Briefly, we divide the query process into \emph{Retrieval} (Entity Extraction, Node Mapping, Path Retrieval) and \emph{Ranking} (Formatting, Ranking and Filtering).
% \zhu{Next, we elaborate on the five main steps of graph query and}
The overall query procedure is illustrated in Figure~\ref{fig:graphQueryPipeline}.

\textbf{Entity Extraction} \zhu{aims to identify various entities in the given speech, treating them as candidate toxic entities for further validation.}
% \textbf{Entity Extraction} faces similar problems as the triplets extracting step in Section \ref{subsec:triplets-extracting}.
\zhu{Given that explicit toxic entities may be absent, we employ the LLM to extract as many relevant entities as possible, including both specific named entities and broader concepts such as race, gender, and religion.}
% In the absence of explicit entities in the query, we employ the LLM to identify the named entities and some other possibly related objects like race, gender, and religion. To extract the related elements, we impart the LLM which kinds of elements should be extracted by providing examples of the elements.
The detailed instruction is shown in Appendix~\ref{appendix:ner-prompt}.

\textbf{Node Mapping} 
%\zhu{faces the challenge that entities extracted from the speech may fail to precisely align with the names of the meta-toxic graph nodes.}
\zyb{faces the challenge that entities extracted from the speech may not exactly match the names of nodes in the meta-toxic knowledge graph.}
To resolve this, we formulate it as a dense retrieval task, 
\zyb{where the most semantically similar node is mapped to each extracted entity based on textual embeddings generated by a pre-trained BERT model.}
%\zhu{mapping the most semantically similar node to each extracted entity based on textual embedding encoded by a BERT model. }
\zyb{At this point, candidate toxic entities are identified within the knowledge graph.}
%\zhu{So far, candidate toxic entities have been located in the knowledge graph.} 
The Faiss library~\cite{douze2024faiss} is applied to optimize retrieval efficiency.
%\gpt{The Faiss library (Douze et al., 2024) is used to enhance retrieval efficiency.}

% \textbf{Node Mapping} stage faces a significant challenge in that the named entities extracted from the queries do not precisely map to the names of Meta Toxic nodes. To address this problem, we formulate it as a dense retrieval task, utilizing the BERT model to encode the entities and the graph nodes to identify the most semantically similar nodes. The retrieval process is optimized through the implementation of the faiss library~\cite{douze2024faiss}.

\textbf{Path Retrieval} 
\zyb{extracts}
% \zhu{evokes} 
\zhu{coherent toxic knowledge contained} 
% \zhu{in}
\zyb{from} \zhu{the graph based on the paths that connect previously mapped nodes through one or multi-hop relations.}
While a straightforward retrieval approach would be to extract all neighbors for each mapped node,
\zhu{this may result in recalling excessive unrelated triplets that could degrade toxicity detection performance, particularly when sensitive entities such as}
% This method often yields excessive unrelated triplets that may compromise the classification accuracy,
``\case{Jew}'', ``\case{Nigger}'', ``\case{Nazi}''. 
\zhu{To mitigate the noise introduced by unnecessary entities or relations,} we enumerate pairwise combinations of \zhu{the mapped nodes and retrieve the shortest path for each node pair. After splitting all the paths into SPO triplets, we take the union of them as the candidate retrieved triplets, reflecting toxicity knowledge potentially related to the text.}

% \textbf{Path Retrieval} stage requires to design a proper search method to retrieve the relevant triplets from the meta-toxic knowledge graph. After mapping the related nodes, while a straightforward approach would be to extract the neighboring nodes of identified entities, this method often yields excessive unrelated triplets that may compromise the classification accuracy, particularly for sensitive entities such as ``Jew'', ``Nigger'', ``Nazi''. Based on the empirical observation that the entities coappearing in the context tend to combine to a shared toxic meaning, we enumerate the pair-wise combinations of the extracted entities and then implement a shortest path search strategy for path retrieval. Then we split each path into SPO triplets and take the union of the retrieved triplets to form the final subgraph.

\textbf{Formatting} transforms the retrieved triplets into candidate knowledge with natural language by concatenating the SPO elements of each triplet, like ``\case{white lives matter is against black lives matter}''. \zhu{After ranking and filtering each remaining knowledge will ultimately be sent to the LLM following this format as part of the prompts.}

\textbf{Ranking and Filtering} is introduced to \zhu{rank the candidate knowledge and discard irrelevant ones for denoising,} as LLMs are sensitive to the prompt content. \zhu{Specifically, we rank the candidate knowledge by sorting the cosine similarity between each candidate’s knowledge and the input speech in descending order. This prioritizes toxicity knowledge that is more relevant to the speech. Additionally, we filter out less related knowledge based on similarity, which are considered as adverse noise.}
% \textbf{Filtering and Ranking} is involved to filter out the irrelevant triplets and rank the relevant triplets for denoising. As the LLMs are sensitive to the context input, it is imperative to implement a robust filtering and ranking mechanism for the retrieved triplets. To keep aligning with the semantics of the query, we prioritize triplets that maintain semantic relevance to the original query content. This is accomplished by computing BERT embeddings for both the query and retrieved triplets, followed by filtering and ranking based on cosine similarity metrics.

\zhu{After querying the meta-toxic knowledge graph, the ultimate toxicity knowledge is retrieved as supplementary information. We then insert the retrieved knowledge into the prompt template utilized for enhancing LLM's ability to detect hatred and toxicity. Two exemplars are designed to inform the LLM on utilizing external toxicity knowledge. The prompting template is presented in Appendix~\ref{appendix:queryprompt}}.

\section{Experiments}\label{sec:experiment}

As outlined in Section~\ref{subsec:data-collection}, we construct three meta-toxic knowledge graphs based on the {HateXplain}, {ToxicSpans}, and {IHC} datasets, respectively, \zyb{with Qwen2.5-14B-Instruct (Qwen)~\cite{qwen2.5}. Then Qwen and Llama3.1-8B-Instruct (Llama)~\cite{llama3modelcard} are employed to generate predictions. We evaluate \emph{\ours} from three perspectives:}
\textbf{(1) Graph \newzyb{Evaluation}}, \newzyb{which provides qualitative and  quantitative analysis on the constructed graphs;}
\textbf{(2) \zhu{Toxicity Prediction}}, where \emph{\ours} is compared to baseline methods under both in-domain and cross-domain settings to evaluate effectiveness and robustness of \emph{\ours};
% by leveraging a meta-toxic knowledge graph built on the {ToxicSpans} dataset and transfer it to the {HateXplain} and {IHC}.
\textbf{(3) Case Studies}, which offer in-depth analyses by guiding LLMs to output reasoning paths, providing insights into how \emph{\ours} enhances interpretability and reasoning abilities. \newzyb{To validate the rationale behind our pipeline design, we have further devised an ablation study, the details of which can be found in Appendix~\ref{appendix:ablation study}}.

\subsection{Graph \newzyb{Evaluation}}

\newzyb{For the constructed KG, to verify whether the triplets contain hate-related semantics and adhere to the required format, we conduct a manual evaluation. }
% For detailed information about the annotations, please refer to the Appendix~\ref{appendix:graph quality}.
To conserve human resources, we sampled 100 triplets from each of the three constructed KGs. Additionally, to mitigate annotators' potential bias towards classifying all triplets as hate-related, we introduce an equal proportion of triplets generated by LLMs that do not contain hate-related information, blending them in a 1:1 ratio.

We assembled a team of 5 annotators who are master's and PhD students, proficient in English, and have some relevant background knowledge. We paid the annotators a wage higher than the minimum wage in their local area. Before annotating, we informed them about the task and ensured that they fully understood and consented to annotate hate speech triplets.

%To conserve human resources, we sampled 100 triplets from each of the three constructed KGs. Additionally, to mitigate annotators' potential bias towards classifying all triplets as hate-related, we introduce an equal proportion of triplets generated by LLMs that do not contain hate-related information, blending them in a 1:1 ratio.
%We assembled a team of 5 annotators who are master's and PhD students, proficient in English, and have some relevant background knowledge. Before annotating, we informed them about the task and ensured that they fully understood and consented to annotate hate speech triplets.
Given the complexity of annotating hate speech triplets, we specifically emphasized three rules: 1) A triplet is considered correct only if it meets the required format and contains hate-related content. 2) When the correctness of a data point was unclear, annotators were instructed to search online first. 3) After completing the annotation, the five annotators should discuss to reach an agreement. The final accuracy of HateXplain, IHC, ToxicSpans KGs is 89\%, 87\% and 82\%, respectively. Most of the triplets are correct, while the main types of errors are format issues and the use of pronouns without specifying the entities they referred to. 

\begin{table}[t]
\caption{Data Summary. $^\dagger$ means that it has been reduced by 21.52\% compared to before the merge; $^\ddagger$ means 3.25\% off.}
% \vspace{-0.3cm}
\label{tab:graphconstructionres}
\centering
\small
\begin{tabular}{ccrr}
\toprule
 & \textbf{Toxic} & \multirow{2}{*}{\textbf{Entities}} & \multirow{2}{*}{\textbf{Triplets}} \\
 & \textbf{Samples} & & \\
\midrule
\textit{IHC} & 7,373 & 20,043 & 25,534 \\
\textit{HateXplain} & 10,273 & 24,442 & 33,276 \\
\textit{ToxicSpans} & 9,905 & 21,667 & 30,347 \\
\midrule
Merged & 27,551 & 51,917$^\dagger$ & 86,350$^\ddagger$ \\
\bottomrule
\vspace{-0cm}
\end{tabular}
\end{table}

To illustrate the scale information of the KGs, we also conducted statistical analysis on the graph.
\zyb{We analyze the graph properties, particularly the number of nodes and relations,} to assess the effectiveness of our data mining approach.
\zyb{Additionally, we examine the merging ratio by integrating meta-toxic knowledge graphs derived from different datasets, demonstrating how the meta-toxic knowledge graph expands.} After merging three datasets, we observe a 21.52\% reduction in the number of entities and a 3.25\% reduction in the number of triplets in the \zyb{merged} graph. This suggests that \zyb{the targets of toxic speech exhibit significant overlap across datasets, supporting the transferability of our approach.}  The detailed quantitative results are presented in Table~\ref{tab:graphconstructionres}.
% , with detailed findings available in the Appendix~\ref{appendix:graph statistics}}. 
To facilitate the understanding of the knowledge graph structure, we illustrate sampled triplets in Appendix~\ref{appendix:example}.

\subsection{Toxicity Prediction}

\zyb{We conduct experiments in both in-domain and cross-domain scenarios, comparing our method with vanilla LLM and the naive RAG-enhanced LLM for toxicity detection.}
\newzyb{Further, to assess 
whether LLMs can outperform fine-tuned small language models , we also include HateBERT~\cite{caselli-etal-2021-hatebert} and BERT~\cite{devlin-etal-2019-bert} as baselines.
% and compared their results with ours. 
It's worth noting that HateBERT requires an initial retraining phase of 18 days and 2 million steps, followed by supervised fine-tuning (SFT) on downstream tasks.}
%between our \textbf{training-free} method and the SOTA \textbf{supervised learning} method based on BERT, we evaluated the performance of the HateBERT~\cite{caselli-etal-2021-hatebert}, which requires an initial pretraining phase of 18 days and 2 million steps, followed by supervised fine-tuning (SFT) on downstream tasks.}
%TODO: Split Toxicity Prediction into in-domain and cross-domain.We conduct experiments on toxicity prediction, comparing our method against vanilla LLM detection and naive RAG-enhanced LLM detection. 
For the vanilla LLM, we directly input the test speech into the LLM and prompt it to provide the prediction. For the naive RAG method, \zyb{we retrieve the top-2 most similar speeches from the training set} as additional knowledge. To ensure a fair comparison, the training set used for the RAG method is the same as the data used to construct our meta-toxic knowledge graph.

\zyb{To apply LLMs for classification,} we follow the evaluation approach from MMLU~\cite{hendryckstest2021}, \zyb{where classification is determined by comparing the logit values of candidate options ``a'' (toxic) and ``b'' (non-toxic).}
% Our method enhances LLMs with a meta-toxic knowledge graph constructed from the same dataset as the test data, including {HateXplain} and {IHC}.
Our evaluation metrics include classification accuracy (Acc.), F1 score (F1), and precision-recall area under the curve (AUC). \zyb{Additionally, we calculate the false positive rate (FPR) to assess how effectively our method reduces false positives.} 

\begin{table*}[t]
    \caption{Results on using {meta-toxic} knowledge graph built from the \textbf{same} dataset, best results are highlighted.}
     \label{tab:indomainres}
    \begin{threeparttable}
        \centering
        \resizebox{0.85\textwidth}{!}{%
            \begin{tabular}{>{\centering\arraybackslash}m{2cm}cccc|ccc|cc} 
            \toprule
            \multicolumn{2}{c}{Backbone Model} & \multicolumn{3}{c|}{Qwen2.5-14B-Instruct} & \multicolumn{3}{c|}{Llama3.1-8B-Instruct} & HateBERT & BERT\\ 
            \cmidrule{1-10}
            \multicolumn{2}{c}{Method} & Vanilla LLM & RAG & \emph{\ours} & Vanilla LLM & RAG & \emph{\ours} & SFT & SFT\\
            \midrule
            \multirow{4}{*}{\textit{HateXplain}} & Acc.$\uparrow$ & 70.95 & 73.13 & {73.39} & 63.36 & {69.59} & 68.87 & 77.96 & \textbf{78.17}\\
            & F1 $\uparrow$& 64.04 & 70.18 & {72.48} & 48.11 & 62.02 & {62.50} & \textbf{81.96} & 81.34\\
            & AUC$\uparrow$ & 76.36 & 83.70 & {84.02} & 79.77 & 80.50 & {82.27} & 88.82 & \textbf{90.44}\\
            & FPR$\downarrow$ & 66.62 & 48.72 & {32.10} & 88.75 & 40.69 &  {38.74} & 31.33 & \textbf{24.94}\\
            \cmidrule{1-10}
            \multirow{4}{*}{\textit{IHC}} & Acc. $\uparrow$& 66.34 & 66.71 & {73.65} & 50.79 & 64.29 & {69.04} & 78.35 & \textbf{78.72}\\
            & F1 $\uparrow$& 66.32 & 66.67 & {69.95} & 48.03 & 64.23& {68.55}& \textbf{71.52} & 70.57\\
            & AUC$\uparrow$& 64.38 & 66.79 & {70.10} & 62.97 & {66.74} &{63.75} & 79.70 & \textbf{79.74} \\
            & FPR$\downarrow$& 48.42 & 43.24 & \textbf{12.32} & 77.63 & 51.58 &{34.23} & 17.49 & 14.19\\
            \bottomrule
            \end{tabular}%
        }
    \end{threeparttable}
\end{table*}
\begin{table*}[t]
    \caption{Results on using {meta-toxic} knowledge graph built from \textbf{another} dataset, best results are highlighted.}
    \label{tab:oodres}
    \begin{threeparttable}
        \centering
        \resizebox{0.85\textwidth}{!}{%
            \begin{tabular}{>{\centering\arraybackslash}m{2cm}cccc|ccc|cc} 
            \toprule
            \multicolumn{2}{c}{Backbone Model} & \multicolumn{3}{c|}{Qwen2.5-14B-Instruct} & \multicolumn{3}{c|}{Llama3.1-8B-Instruct} & HateBERT & BERT\\ 
            \cmidrule{1-10}
            \multicolumn{2}{c}{Method} & Vanilla LLM & RAG & \emph{\ours} & Vanilla LLM & RAG & \emph{\ours} & SFT & SFT\\
            \midrule
            \multirow{4}{*}{\textit{HateXplain}} & Acc.$\uparrow$ & 70.95 & 71.21 & \textbf{73.28} & 63.36 & 65.96 & {68.24} & 59.30 & 57.38\\
            & F1$\uparrow$ & 64.04 & 64.97 & \textbf{72.38} & 48.11 & 54.32 & {61.63} & 69.11 & 45.78\\
            & AUC$\uparrow$& 76.36 & 81.29 & \textbf{83.80} & 79.77 & 81.50 & {81.67} & 72.63 & 58.57\\
            & FPR$\downarrow$& 66.62 &64.32 & \textbf{32.10} & 88.75 & 80.95 & {67.13} & 66.11 & 86.32\\
            \cmidrule{1-10}
            \multirow{4}{*}{\textit{IHC}} & Acc. $\uparrow$& 66.34 & 63.36 & \textbf{73.42} & 50.79 & 52.84 & {69.04} & 45.86 & 40.36\\
            & F1 $\uparrow$& 66.32 & 63.34 & \textbf{69.42} & 48.03 & 51.01 & {68.85} & 45.06 & 37.66\\
            & AUC$\uparrow$& 64.38 & 67.21 & \textbf{69.78} & 62.97 & 61.15 & {63.63} & 39.88 & 33.27\\
            & FPR$\downarrow$& 48.42 & 43.24 & \textbf{11.64} & 77.63 & 72.97 & 34.23 & 61.86 & 84.23\\
            \bottomrule
            \end{tabular}%
        }
    \end{threeparttable}
\end{table*}

\subsubsection{In-domain Setting}\label{sec:ids}
For the in-domain setting, the graph construction and toxicity detection are performed on the training and test sets of the same dataset, respectively.

As illustrated in Table~\ref{tab:indomainres}, \emph{\ours} demonstrates performance improvements across the HateXplain and IHC datasets with two backbone models, \zyb{consistently outperforming baseline methods.}
The results reveal key insights into the effectiveness of our approach.
For HateXplain, where toxic semantics are relatively explicit, the related knowledge is easier to retrieve by naive RAG, improving the performance to some extent. However, \emph{\ours} outperforms naive RAG, primarily due to the more concise and relevant knowledge provided by KG.

In contrast, for IHC, where toxic semantics are more implicit, overall performance is lower compared to HateXplain. \zyb{Using Qwen, while naive RAG yields only marginal improvement, \emph{\ours} achieves a remarkable performance boost.} This discrepancy may stem from the fact that documents retrieved by naive RAG often contain excessive unrelated information. {In contrast,} \emph{\ours}, through the combined contributions of knowledge graph construction and query on the graph, not only retrieves the most related knowledge but also organizes it into concise triplets, significantly improving the model's understanding of the provided knowledge.
When using the less powerful Llama for detection, \emph{\ours} {\zyb{achieves performance slightly better than naive RAG, reflecting the advantage of its more effective knowledge integration, even with limited model capacity.}
}{\zyb{In brief, our method effectively enhances the performance for in-domain scenario.}}

\newzyb{Compared to HateBERT and fine-tuned BERT, 
% it is
surprisingly,
we 
% are surprised to 
find that \textbf{a simple BERT-based model, after training, can outperform a 14B LLM under in-domain scenarios}. The key distinction lies in the way of capturing toxicity: our method explicitly presents hate-related information with KG triplets, whereas HateBERT implicitly recognizes toxicity patterns by SFT. 
We speculate that small models' performance advantage attributes to supervised learning, which enables them to memorize toxicity knowledge within the training data~\cite{chu2025sftmemorizesrlgeneralizes}.
While HateBERT performs well 
% We speculate the performance advantage of small models to supervised learning, which enables them to memorize toxicity knowledge from the training data~\cite{chu2025sftmemorizesrlgeneralizes}.
% Training with representations is more effective for handling static datasets with the same distribution, 
on static datasets, toxic detection requires rapid updates in fast-evolving social media environments. This makes our approach, which constructs a knowledge graph that is easier to edit and expand, more suitable for such tasks. Therefore, we conduct experiments in a cross-domain setting as discussed below.}

\begin{figure}[t]
    \centering\resizebox{\linewidth}{!}
    {\includegraphics[width=\linewidth]{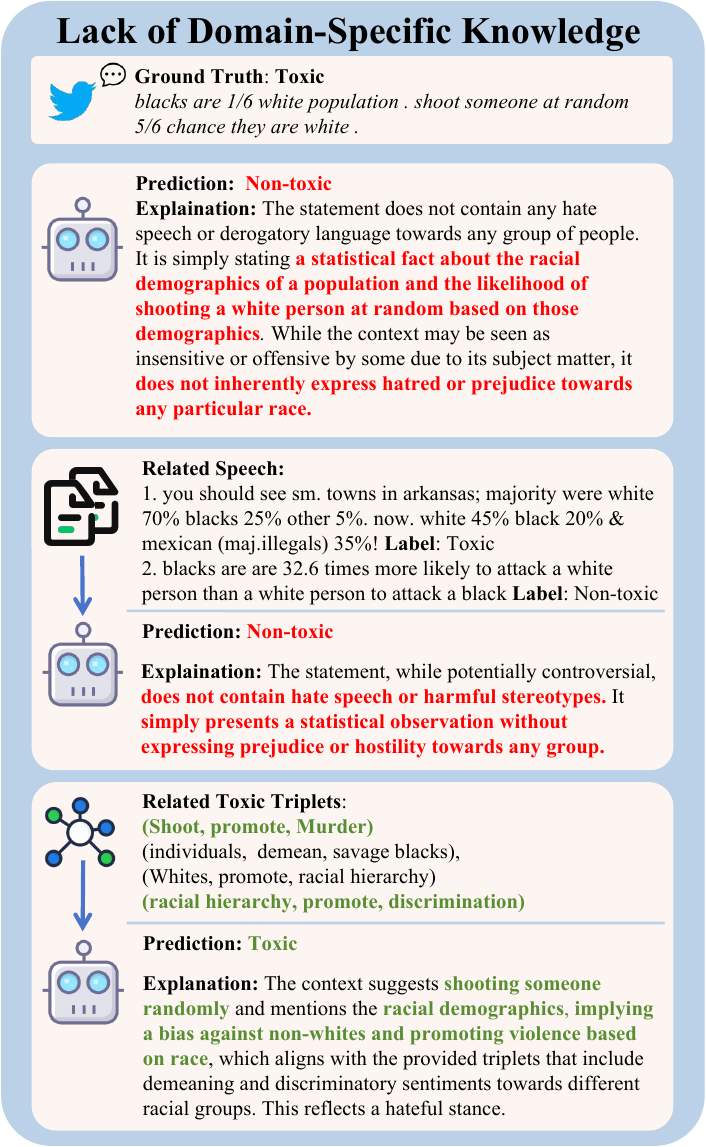}}
    \caption{Example of false negative misjudgment caused by lack of domain-specific knowledge. \zyb{{From top to bottom, each block shows the ground truth label and context, followed by the predictions and reasoning of the vanilla LLM, naive RAG, and \emph{\ours}.}}}
    \label{fig:knowledge_case}
\end{figure}

\begin{figure}[t]
    \centering\resizebox{\linewidth}{!}
    {\includegraphics[width=\linewidth]{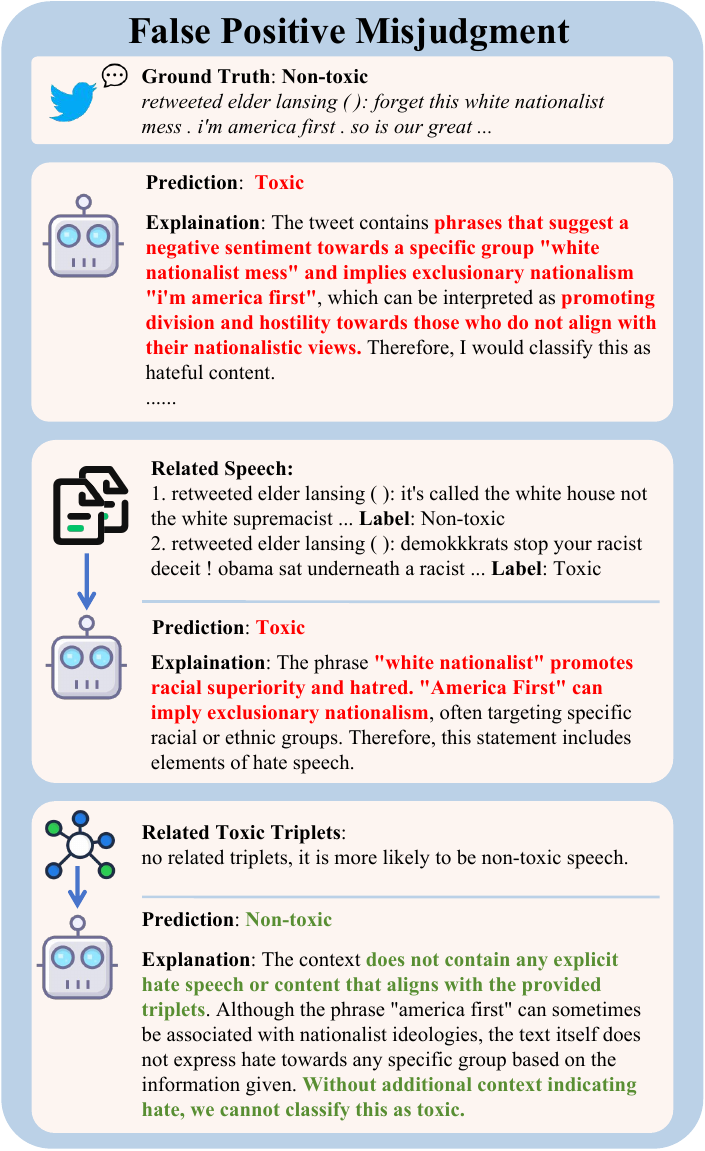}}
    \caption{Example of false positive misjudgment caused by LLMs' extreme sensitivity towards certain phrases. The blocks are distributed the same as Figure~\ref{fig:knowledge_case}.}
    \label{fig:fp_case}
\end{figure}

\subsubsection{Cross-domain Setting}
For the cross-domain setting, the knowledge graph is constructed using the training set of ToxicSpans, while detection is performed on test sets of HateXplain and IHC. Correspondingly, naive RAG is only allowed to retrieve documents from Toxicspans. 

The experimental results are presented in Table~\ref{tab:oodres}. Notably, we observe a performance drop with naive RAG on IHC and negligible improvement on HateXplain. \zhu{This can be attributed to the different facets of toxic semantics between ToxicSpans and test datasets.}
\newzyb{As analyzed in Section~\ref{sec:ids}, the text representation-based method, HateBERT and BERT, struggles in OOD scenarios, resulting in a significant performance drop.
HateBERT performs slightly better than BERT, likely due to its extensive retraining on a relatively large toxicity corpora, RAL-E dataset.
In contrast, \emph{\ours} still maintains a notable improvement in performance, because by incorporating external accurate toxicity knowledge, LLMs can uncover the underlying, invariant toxicity essence concealed within the diverse expressions of toxic content, therefore enhancing generalization.} 

% It is worth noting that, in all scenarios, our method not only maintains superior performance but also achieves the lowest false positive rate, effectively mitigating the risk of infringing on freedom of speech caused by false positives.

\newzyb{It is worth noting that our method outperforms other LLM-based approaches across all scenarios, achieving the lowest or near-lowest false positive rate, effectively mitigating the risk of infringing on freedom of speech caused by false positives.}

\subsection{Case Study}

To further analyze how \emph{\ours} incorporates domain-specific knowledge and mitigates false positive errors, we present two case studies 
with the reasoning explanations provided by the LLM.
% where the LLM is instructed to provide reasoning explanations.

As shown in Figure~\ref{fig:knowledge_case}, the original speech implicitly promotes toxicity by stirring up antagonism between blacks and whites. The word ``\case{shoot}'' also emphasizes the racial disparities. The vanilla LLM incorrectly focuses on numbers like ``\case{1/6}'' and ``\case{5/6}'', which misleads the LLM to interpret the speech as a statistical analysis rather than a toxic trigger.
% Enhanced with naive RAG, the retrieved related speeches emphasize statistics like ``70\%'', and ``32.6 times'', \zyb{as naive RAG fails to capture the most crucial semantics.} \zyb{This further misguides the LLM, leading it to incorrectly interpret the original speech as a statistical statement, resulting in a misjudgment with overly high confidence.}
\newzyb{When enhancing with naive RAG, the retrieved relevant texts contain statistics like ``\case{70\%}'', and ``\case{32.6 times}'', indicating that the retriever fails to understand the intended meaning of the text. This further deepened the model's misunderstanding, leading to misjudgments.}
%to misunderstand the original speech is indeed a statistical statement, resulting in misjudgment and extremely high confidence.
On the contrary, \emph{\ours} correctly identifies pivotal elements like ``\case{shoot}'',  and ``\case{racial hierarchy}'', guiding the model to correctly interpret the speech as promoting race-based violence and predict correctly.

As shown in Figure~\ref{fig:fp_case}, the original context emphasizes unity by stating ``\case{I'm America first,}'' \zyb{while suggesting} ignoring toxic content presented by ``\case{white nationalist mess}''. However, the vanilla LLM misclassifies it as toxic because it \zyb{overly} focuses on phrases like ``\case{America First}'' and ``\case{white nationalist mess}'', incorrectly interpreting them as indicators of toxicity.
When enhanced with naive RAG, the LLM retrieves two related retweets for the same post. 
However, since both retweets focus on racial issues, even though they express different opinions, they fail to shift the LLM's focus away from racial attributes, resulting in a false positive misjudgment. In contrast, \emph{\ours} effectively filters the retrieved triplets, \zyb{directing} the model to follow the prompt instructions that ``\case{when there are no related triplets, the text is more likely to be non-toxic}'', which significantly reduces the false positive misjudgment.

% Given the absence of ground truth for toxicity reasoning, we conducted comprehensive case studies and employed an LLM-as-Judge approach to assess our approach's effectiveness. We randomly sampled 20 instances from the {IHC} test set for evaluation. In the LLM-as-Judge framework, we presented GPT-4 with the original text alongside reasoning outputs from both the vanilla LLM and our method, instructing it to identify the more cogent explanation. To ensure fair comparison and mitigate potential verbosity bias~\cite{saito2023verbosity}, we standardized the maximum token length to 256 across all methods.

% The evaluation results, following manual verification by the authors, revealed that our proposed method and the baseline approach demonstrated equivalent performance in one case. Our approach demonstrated superior reasoning in 11 cases, while the baseline showed marginal advantages in the remaining 8 instances.

% Representative examples are illustrated in Figures \ref{fig:case1} and \ref{fig:case2}. In this visualization, green text highlights reasoning components that the judge deemed strongly indicative of the underlying causes, while yellow text denotes explanations considered less satisfactory. 
% We can find that the meta-toxic knowledge graph effectively augments LLMs' reasoning capabilities, producing more precise and well-grounded explanations. However, we observed instances where the judge deemed certain explanations overly specific, which may lead to mistakes.

\section{Conclusion}

% In this paper, we designed a three-step pipeline to extract accurate toxic triplets and constructed a domain-specific meta-toxic knowledge graph for toxic speech detection.
% We will open-source the knowledge graph upon paper acceptance.
% Through the construction of the meta-toxic knowledge graph, we successfully extracted the meta-level toxic information from existing datasets.
In this paper, we proposed a novel method called \emph{\ours} to address both false positive and false negative misjudgments caused by the lack of domain-specific knowledge and LLMs' extreme sensitivity. First, we leverage LLMs to extract toxic content through a three-step pipeline, which builds the meta-toxic knowledge graph. Next, we query the graph with retrieval and ranking processes to provide additional, relevant toxic knowledge. Extensive experiments and detailed case studies across various datasets show that \emph{\ours} significantly lowers the false positive rate while improving overall toxicity detection performance in OOD scenarios. In addition, under in-domain scenarios, we surprisingly find that small language models are more competent.

\section*{Acknowledgement}

This work was supported by Shanghai ``Science and Technology Innovation Action Plan'' Project under Grant No.23511100700 and National Natural Science Foundation of China under Grant 42375146.

% \clearpage
\section*{Limitation}

Our study still has several limitations. First, due to the computational constraint, we did not conduct experiments with larger LLMs, suggesting that the full potential of our method remains to be further explored. Future research could leverage more powerful LLMs to construct even more comprehensive meta-toxic knowledge graphs. Additionally, our method is currently limited to binary classification in English-language scenarios. Future work can extend our approach to multi-modal, multi-lingual, and multi-cultural tasks, thereby broadening its applicability across diverse contexts.

% Additionally, our method heavily relies on LLMs, which may introduce relatively high computational overhead. How to achieve a more lightweight version of the method while maintaining performance remains an open issue for further discussion. 
\section*{Ethical Statement}

Our research focuses on toxic speech detection, primarily addressing two key issues faced by LLMs in this task: false negatives caused by a lack of domain-specific knowledge, and false positives resulting from excessive sensitivity to certain content. Our primary goal is to facilitate more accurate detection of toxicity, thereby alleviating the issue of false positive misjudgments and protecting freedom of speech, which aims to contribute to the creation of a more harmonious online environment.
We construct a meta-toxic knowledge graph filled with toxic content. However, there is also a risk of potential misuse of our proposed method. Specifically, our methods might be misused to generate toxic speech. It is essential to emphasize that our work focuses on detecting toxic speech while safeguarding freedom of speech, instead of censorship.
The datasets we used are all existing open-source datasets, aligning with their intention for scientific research. We also adhered to the MIT license for HateXplain and IHC datasets, and the CC0-1.0 license for the ToxicSpans dataset.
% \input{_txt/6.futurework}

\appendix
\clearpage
\section{Prompts}

\subsection{Rationale Reasoning Prompt}\label{appendix:reasoning-prompt}
\begin{figure}[H]
\centering\includegraphics[width=0.95\textwidth]{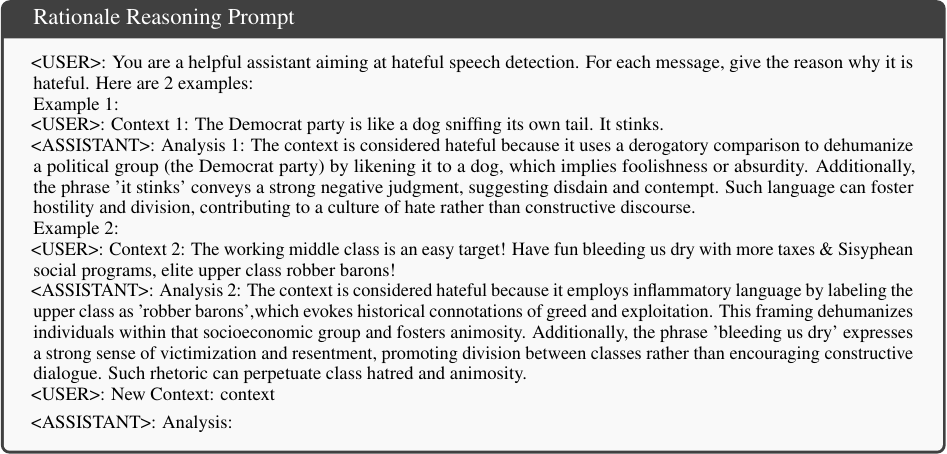}
% \captionof{figure}{Rationale Reasoning Prompt}
\label{rationale}
\end{figure}

\subsection{Triplet Extraction Prompt}\label{appendix:extracting-prompt}

\begin{figure}[H]

\centering\includegraphics[width=0.95\textwidth]{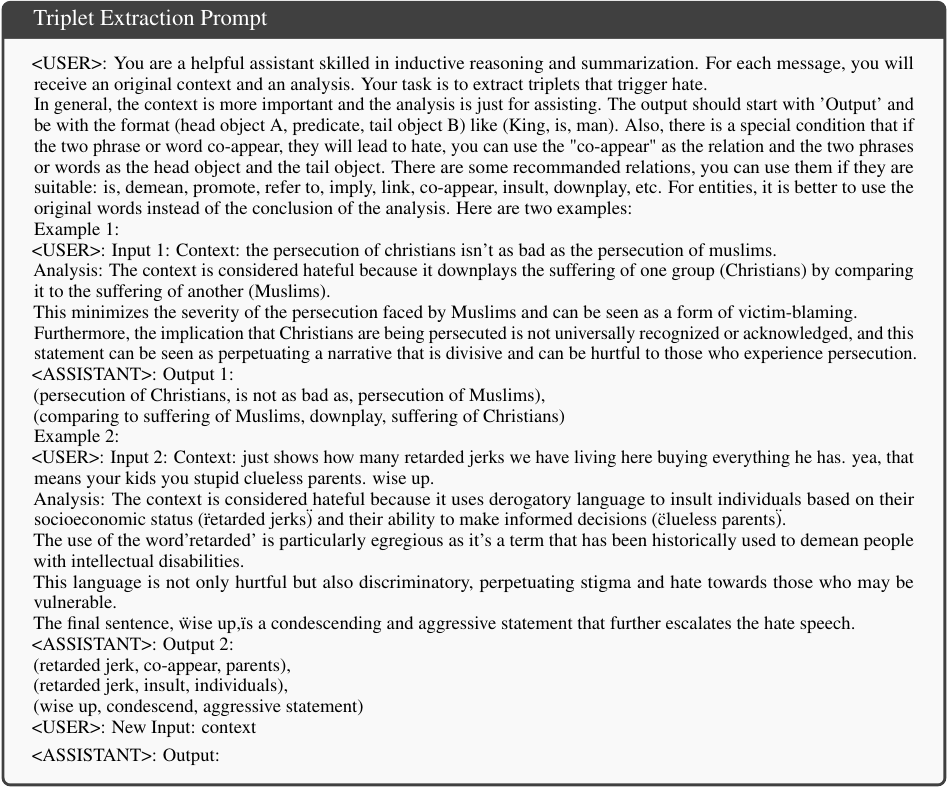}

% \captionof{figure}{Triplet Extraction Prompt}
\label{triplets}
\end{figure}
\clearpage

\subsection{Self-checking Prompt}\label{appendix:filtering-prompt}
\begin{figure}[H]
\centering\includegraphics[width=0.95\textwidth]{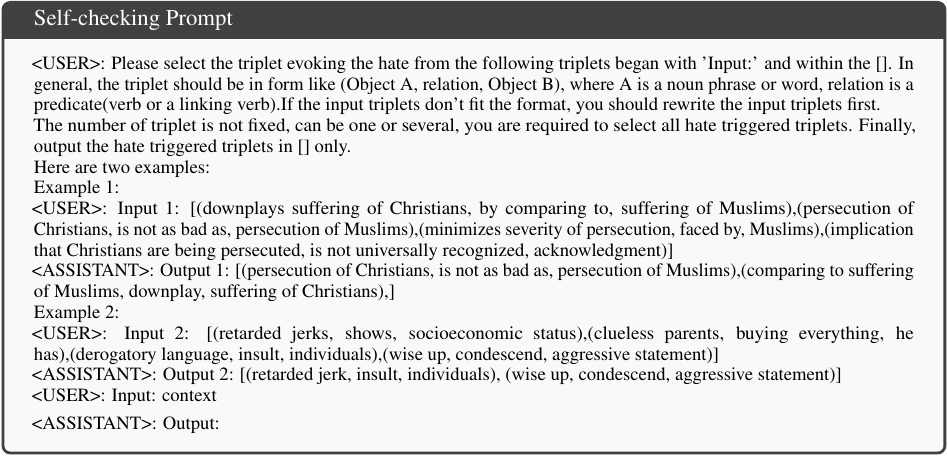}\label{selfchecking}    
% \captionof{figure}{Self-checking Prompt}
\end{figure}

\subsection{NER Prompt}\label{appendix:ner-prompt}

\begin{figure}[H]
\centering\includegraphics[width=0.95\textwidth]{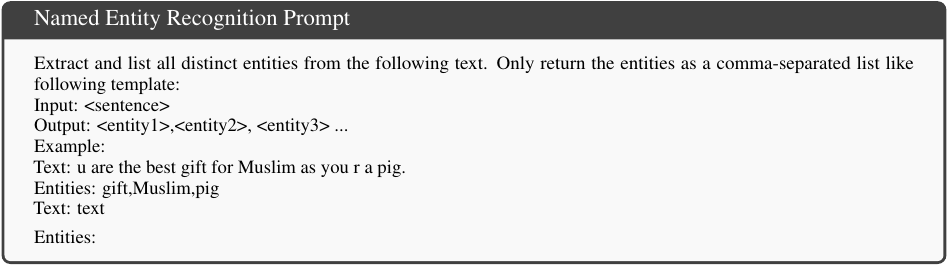}
% \captionof{figure}{NER Prompt}
\label{ner}    
\end{figure}

\subsection{Query Prompt}\label{appendix:queryprompt}

\begin{figure}[H]
\centering\includegraphics[width=0.95\textwidth]{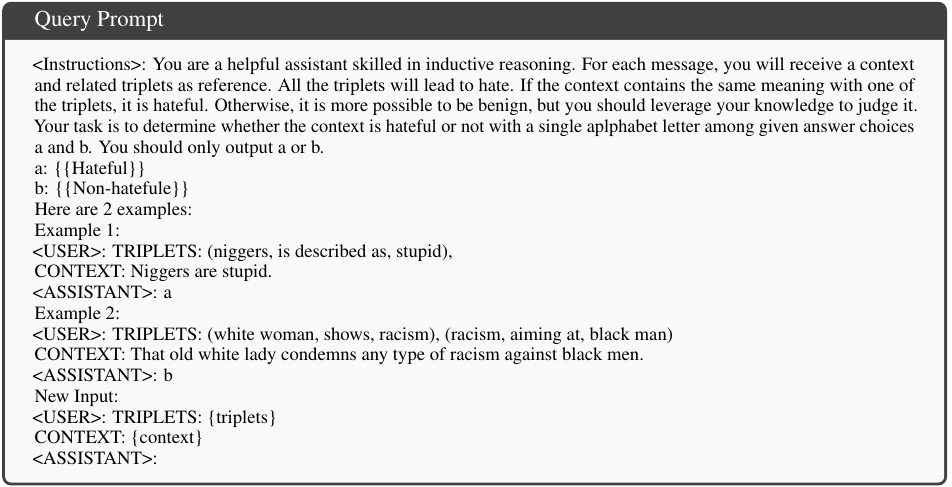}
\label{query}
\end{figure}
\clearpage

% \clearpage

% \begin{table*}[ht]
% \centering
% \phantomsection
% \begin{tcolorbox}[colback=white!95!gray,colframe=gray!50!black,rounded corners,label={scale-depression}, title={Query Prompt}]
% \begin{lstlisting}
% <Instructions>: You are a helpful assistant skilled in inductive reasoning. For each message, you will receive a context and related triplets as reference. All the triplets will lead to hate. If the context contains the same meaning with one of the triplets, it is hateful. Otherwise, it is more possible to be benign, but you should leverage your knowledge to judge it. Your task is to determine whether the context is hateful or not with a single aplphabet letter among given answer choices a and b. You should only output a or b.

% a: \{\{Hateful\}\}

% b: \{\{Non-hatefule\}\}

% Here are 2 examples:

% Example 1:

% <USER>: TRIPLETS: (niggers, is described as, stupid), 

% CONTEXT: Niggers are stupid.

% <ASSISTANT>: a

% Example 2:

% <USER>: TRIPLETS: (white woman, shows, racism), (racism, aiming at, black man)

% CONTEXT: That old white lady condemns any type of racism against black men.

% <ASSISTANT>: b

% New Input:

% <USER>: TRIPLETS: \{triplets\}

% CONTEXT: \{context\}

% <ASSISTANT>: 
% \end{lstlisting}
% \end{tcolorbox}
% \label{tab:generate_prompt}
% \end{table*}

\section{Ablation Study}\label{appendix:ablation study}

To validate the effectiveness and rationality of each step in our retrieval pipeline, we designed an ablation study. We selected Qwen-14B-Instruct as the backbone model and conducted the experiment on the HateXplain dataset. Specifically, we compared the settings without rerank \& filter and those replacing shortest path search with 1-hop neighbor search. The results are presented in Table~\ref{tab:abstudy}.

By comparing the results of the first and second rows, we find that rank \& filter significantly improves performance. Similarly, comparing the second and third rows shows that shortest path search outperforms 1-hop neighbor search. This may be because the elements involved in shortest path search are more relevant, reducing input noise.

\begin{table}[h]
\caption{The ablation results on the HateXplain dataset based on Qwen2.5-14B-Instruct.}
\label{tab:abstudy}
\centering
\small
\begin{tabular}{ccc}
\toprule
 method & Acc. & F1 \\
\bottomrule
\textbf{MetaTox}(w/ rank \& filter, shortest path) & \textbf{73.39} & \textbf{72.48} \\
w/o rank \& filter, w/ shortest path & 70.17 & 71.73 \\
w/o rank \& filter, w/ 1-hop subgraph & 65.38 & 69.18 \\
\bottomrule
\end{tabular}
\end{table}

\section{Examples in Knowledge Graph}\label{appendix:example}

\begin{figure}[H]
\centering\includegraphics[width=1.1\linewidth]{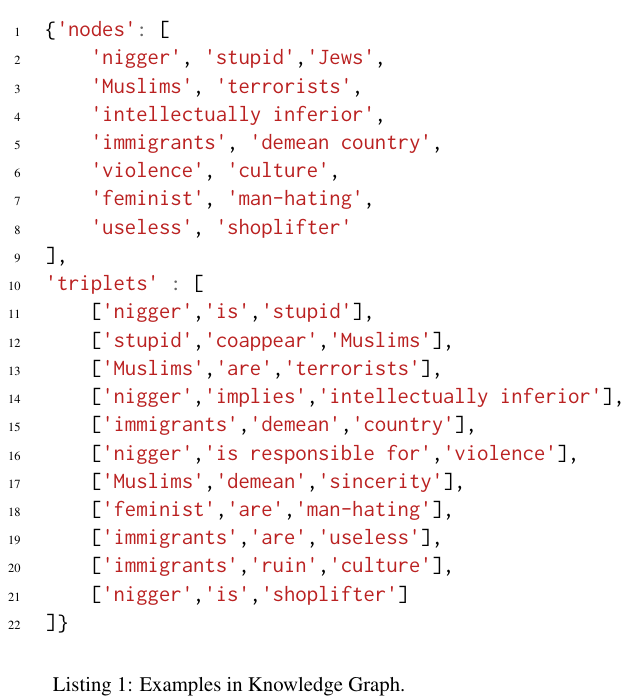}    
% \captionof{figure}{Self-checking Prompt}
\end{figure}

\end{document}